\documentclass[letterpaper]{article} 
\usepackage{aaai25}  
\usepackage{times}  
\usepackage{helvet}  
\usepackage{courier}  
\usepackage[hyphens]{url}  
\usepackage{graphicx} 
\usepackage{natbib}  
\usepackage{caption} 
\usepackage{algorithm}
\usepackage{algorithmic}
\usepackage{verbatim}
\usepackage{multirow}
\usepackage{subfigure}
\usepackage{amsmath}
\usepackage{amsthm}
\usepackage{amssymb}
\usepackage{mathrsfs}
\usepackage{graphicx}
\usepackage{adjustbox}
\usepackage{booktabs}
\usepackage{xcolor}

\newtheorem{theorem}{Theorem}
\usepackage{newfloat}
\usepackage{listings}
\DeclareCaptionStyle{ruled}{labelfont=normalfont,labelsep=colon,strut=off} 
\floatstyle{ruled}
\newfloat{listing}{tb}{lst}{}
\floatname{listing}{Listing}
\title{A Simple Graph Contrastive Learning Framework for Short Text Classification}
\author {
    Yonghao Liu\textsuperscript{\rm 1},
    Fausto Giunchiglia\textsuperscript{\rm 2},
    Lan Huang\textsuperscript{\rm 1},
    Ximing Li\textsuperscript{\rm 1}, \\
    Xiaoyue Feng\textsuperscript{\rm 1}\thanks{Corresponding Author.},
    Renchu Guan\textsuperscript{\rm 1}\footnotemark[1]
}
\affiliations {
   \textsuperscript{\rm 1}Key Laboratory of Symbolic Computation and Knowledge Engineering of the Ministry \\ of Education, College of Computer Science and Technology, Jilin University \\
    \textsuperscript{\rm 2}University of Trento \\
    yonghao20@mails.jlu.edu.cn, fausto.giunchiglia@unitn.it, liximing86@gmail.com, \\ \{huanglan, fengxy, guanrenchu\}@jlu.edu.cn 
}

\usepackage{bibentry}

\begin{document}

\maketitle

\begin{abstract}
Short text classification has gained significant attention in the information age due to its prevalence and real-world applications. Recent advancements in graph learning combined with contrastive learning have shown promising results in addressing the challenges of semantic sparsity and limited labeled data in short text classification. However, existing models have certain limitations. They rely on explicit data augmentation techniques to generate contrastive views, resulting in semantic corruption and noise. Additionally, these models only focus on learning the intrinsic consistency between the generated views, neglecting valuable discriminative information from other potential views.
To address these issues, we propose a \textbf{Sim}ple graph contrastive learning framework for \textbf{S}hort \textbf{T}ext \textbf{C}lassification (\textbf{SimSTC}). Our approach involves performing graph learning on multiple text-related component graphs to obtain multi-view text embeddings. Subsequently, we directly apply contrastive learning on these embeddings. Notably, our method eliminates the need for data augmentation operations to generate contrastive views while still leveraging the benefits of multi-view contrastive learning. Despite its simplicity, our model achieves outstanding performance, surpassing large language models on various datasets. 
\end{abstract}

\section{Introduction}
In the era of information, we are surrounded by large amounts of short texts, such as tweets, news headlines, and product reviews. Efficiently extracting valuable information from short texts is not easy. Compared to regular texts, \textit{on the one hand}, short texts typically have limited contextual information and serious semantic sparsity issues \cite{wang2021hierarchical}. They may only contain a few words, which increases the difficulty of correctly understanding their semantics \cite{phan2008learning}. \textit{On the other hand}, unlabeled short texts far outnumber labeled ones, leading to a severe label scarcity problems \cite{linmei2019heterogeneous}. Short text classification (STC), as a fundamental task of natural language processing, has wide applications in real life, such as sentiment analysis \cite{liu2021deep, li2024simple}, social media analysis \cite{wu2014opinionflow, liu2021vpalg}, and intent recognition \cite{chen2019deep, liu2023time}. Compared to regular text classification, STC has sparked significant research enthusiasm among researchers. Recently, some studies \cite{liu2024improved} have explored the combination of graph neural networks (GNNs) and contrastive learning (CL) to address STC tasks, aiming to utilize the strengths of both to learn informative text features, with impressive performance. 
These models generally start by constructing a corpus-level heterogeneous graph and perform GNNs on it to obtain text embeddings. Subsequently, they leverage CL techniques to aim to maximize the utilization of valuable information contained in numerous unlabeled short texts, which can facilitate the model to learn discriminative text representations.

Despite their fruitful success, there are some limitations that hinder further performance improvements. \textbf{First}, these models almost need to perform explicit data augmentation to generate contrastive views of short texts for CL. However, \textit{on the one hand}, the optimal data augmentation configuration largely depends on extensive empirical trials, such as selecting combinations of data augmentations and the hyperparameter settings \cite{li2023augmentation}. Additionally, the increasing number of hyperparameters for data augmentation exponentially expands the search space for data augmentation approaches, undoubtedly adding significant difficulty. Therefore, finding an optimal data augmentation configuration requires numerous time and computational resources. \textit{On the other hand}, an inappropriate data augmentation can even have a detrimental effect. Specifically, these models either perform random word insertion or deletion operations on the original text to obtain augmented text \cite{chen2022contrastnet}, or they perform perturbation operations on nodes or edges of the constructed text graph to obtain augmented one \cite{su2022contrastive}. However, in either way, it inevitably damages the semantic information of the text and introduces noise. For example, if we randomly insert a negation word into a short text like ``this is a good movie'', it becomes ``this is a not good movie''. This completely changes the original semantics and even turning it into a negative sample with dissimilar semantics, which can have a negative impact on the subsequent CL, leading to suboptimal results. Hence, a natural question arises: is there an elegant method to generate contrasting views of text without performing data augmentation?

\textbf{Second}, typically, 
the information provided by different views complements each other, allowing the model to capture more comprehensive information for learning data embeddings, which may be helpful for downstream tasks \cite{tian2020contrastive}. However, existing models only require two views to perform CL for learning the intrinsic consistency of the data, while ignoring the valuable information contained in potential views, which inevitably impairs the model performance. 

To address the issues mentioned above, we propose a simple graph contrastive learning framework named SimSTC for STC. Specifically, we first build three text-related multi-view component graphs: word graph,  part-of-speech (POS) graph, and entity graph, and perform graph learning on them to obtain text embeddings from different perspectives. These component graphs can, on the one hand, alleviate the semantic sparsity issue of short texts by providing richer semantic and syntactic information, and on the other hand, offer different interpretations of the target short text from various perspectives. In other words, the obtained text embeddings related to different component graphs naturally form the required contrastive views without the need for any data augmentation operations. Furthermore, to capture useful information between different views, we perform CL on all three involved views naturally to effectively address the issue of missing information. Interestingly, we find that even without using sophisticated techniques, our proposed simple framework still achieves impressive performance, consistently outperforming previous competitive models and even surpassing recent popular large language models on several benchmark datasets. We also provide relevant theoretical analysis, which indicates that our approach can enhance the mutual information between different views, thereby improving the model performance.

In summary, our key contributions are listed below.

\noindent $\bullet$ We develop a simple graph contrastive learning framework, namely SimSTC, for STC tasks. It does not require data augmentation and directly uses the three views formed during the construction of component graphs as contrastive views. Compared to regular dual-view CL, it naturally handles the issue of information loss caused by missing views. 

\noindent $\bullet$ We provide rigorous theoretical analysis from the perspective of mutual information behind the simple empirical framework to delve into why it works, thus offering theoretical guarantees.

\noindent $\bullet$ We conduct extensive experiments on the evaluated datasets and the results demonstrate the effectiveness of our proposed SimSTC, which even considerably outperforms large language models on several datasets. 
\section{Related Work}
\subsection{Short Text Classification}
Short texts suffer from severe semantic sparsity issues, as they typically contain only a few words, greatly increasing the difficulty of understanding them correctly \cite{wang2017combining}. Additionally, since the vast majority of short texts in real life are unlabeled, this leads to a significant problem of label scarcity \cite{linmei2019heterogeneous}. The two aforementioned issues make STC even more challenging. Efforts have been made to extract additional semantic and syntactic information from internal corpora. These include mining latent topics \cite{linmei2019heterogeneous} and syntactic dependencies \cite{liu2020tensor}, as well as incorporating relevant entity information from external knowledge graphs \cite{chen2019deep}. These approaches aim to enhance the information contained in short texts, resulting in notable improvements. 
However, they only partially alleviate the semantic sparsity issue and do not substantially address the label scarcity problem. Thus, some studies \cite{yang2021hgat,wang2021hierarchical,liu2025boost} propose constructing short texts into graphs based on co-occurrence of words or phrases in the corpus, and then using the message-passing mechanism of graph neural networks (GNNs) for label propagation \cite{liu2022few, liu2023global}, effectively alleviating the label sparsity issue. Recently, some work \cite{su2022contrastive, liu2024improved} has aimed to integrate the advantages of CL, which enables learning discriminative representations without the need for labels, with graph learning. This integration aims to simultaneously leverage the strengths of both techniques to extract self-supervised signals present in massive unlabeled data, which can then be used to assist in STC tasks. 

\subsection{Contrastive Learning}
CL has been widely practiced in various fields, from computer vision \cite{chen2020simple,he2020momentum}, natural language processing \cite{giorgi2021declutr,gao2021simcse}, to graph learning \cite{liu2024simple, liu2024meta}. It has been demonstrated to extract expressive representations that are comparable to those obtained through supervised learning, even in the absence of labeled data, for downstream task analysis. 
The core idea of CL is to minimize the distance between positive pairs formed by an original sample and its augmented counterpart, while maximizing the distance from negative pairs formed by other samples. Some pioneering work \cite{wu2018unsupervised,oord2018representation} treats each instance as a unique class and performs instance discrimination tasks in an unsupervised manner. Additionally, other work \cite{tian2020contrastive} has achieved impressive performance by performing CL through multi-view image construction, learning expressive image encodings in the absence of labels. Expanding into the field of natural language processing, 
recent proposed models \cite{chen2022contrastnet,pan2022improved} have successfully incorporated CL to address text classification tasks, yielding satisfactory performance. These studies focus on how to generate suitable positive and negative samples required for CL in text. In other words, they both require data augmentation operations to obtain contrastive views, and their performance heavily depends on the parameters chosen for the data augmentation configuration. Moreover, they only perform CL on two views, neglecting valuable information contained in other potential views, which further limits the model's expressive capacity.
\section{Preliminary}
In this section, we introduce relevant techniques to be used as background knowledge. Classic GNNs \cite{kipf2016semi,velivckovic2017graph,hamilton2017inductive} employ a message-passing mechanism, iteratively updating their node features by aggregating information from neighboring nodes. Here, we adopt a pioneering GCN \cite{kipf2016semi} for simplicity and efficiency, which can be defined formally as follows.
\begin{equation}
    \label{gnn}
    \mathbf{H}^{(\ell+1)} = \sigma(\hat{\mathbf{D}}^{-\frac{1}{2}} \hat{\mathbf{A}}\hat{\mathbf{D}}^{-\frac{1}{2}}\mathbf{H}^{(\ell)}\mathbf{W}^{(\ell)}),
\end{equation}
where $\hat{\mathbf{A}}=\mathbf{A}+\mathbf{I}$ symbols an adjacency matrix with added self-loops, and $\hat{\mathbf{D}}_{ii}=\sum_j\hat{\mathbf{A}}_{ij}$ denotes the diagonal degree matrix. $\mathbf{H}^{(\ell)}$ denotes the $\ell$-th output node embedding and $\mathbf{H}^{(0)}=\mathbf{X}$ is the initial node embedding. $\sigma(\cdot)$ is an activation function such as ReLU and $\mathbf{W}^{(\ell)}$ represents the trainable matrix. By iteratively performing Eq.\ref{gnn}, we can obtain the updated node embeddings for downstream tasks.
\section{Method}
\begin{figure*}
    \centering
    \includegraphics[width=0.9\textwidth]{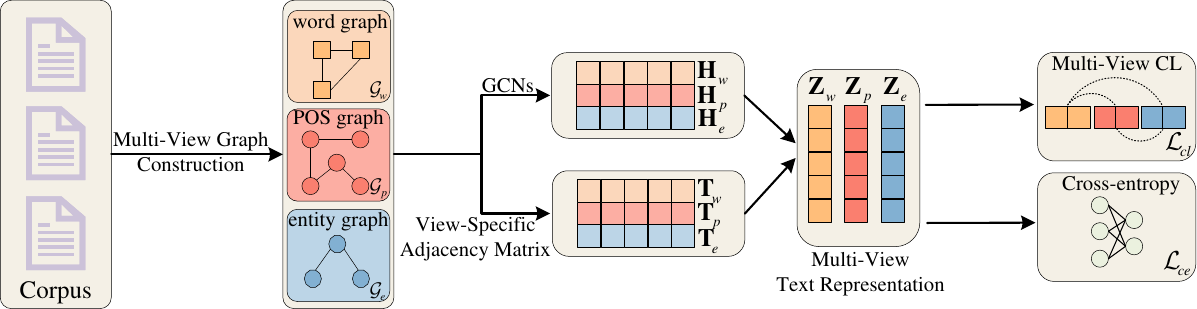}
    \caption{The overall architecture of SimSTC. First, we perform multi-view graph construction on the corpus to obtain three text-related graphs $\mathcal{G}_w, \mathcal{G}_p$ and $\mathcal{G}_e$. Then, we apply GCNs to each constructed graph to obtain high-quality node embeddings $\mathbf{H}_w, \mathbf{H}_p$, and $\mathbf{H}_e$. Meanwhile, we acquire view-specific adjacency matrices $\mathbf{T}_w, \mathbf{T}_p$, and $\mathbf{T}_e$. After obtaining $\mathbf{H}_\pi$ and $\mathbf{T}_\pi$, we perform information aggregation operations on them to obtain multi-view text embeddings $\mathbf{Z}_w, \mathbf{Z}_p$, and $\mathbf{Z}_e$. Finally, we apply multi-view CL and cross-entropy on the obtained text embeddings.}
    \label{flowchart}
\end{figure*}

In this section, we will elaborate on each component of the proposed framework. Specifically, it mainly consists of three parts: component graph construction, multi-view text representation learning, and multi-view contrastive learning. To facilitate better understanding, we present the overall flowchart in Fig. \ref{flowchart}.

\subsection{Multi-View Graph Construction}
We can leverage the rich and effective semantic and synthetic information contained within the given specific corpus, along with the auxiliary information contained in the externally related knowledge base, to compensate for the understanding difficulties caused by the limited context of the target short text. Therefore, we propose to construct text-related graphs under three different views, including word graph, POS graph, and entity graph. We provide the specific process of graph construction in the following.

The word graph $\mathcal{G}_w=\{\mathcal{V}_w,\mathbf{X}_w,\mathbf{A}_w\}$, comprising the lexical units within the given short text, facilitates the provision of comprehensive semantic information. $\mathcal{V}_w$ represents the word set and $\mathbf{X}_w \in \mathbb{R}^{|\mathcal{V}_w|\times d_w}$ represents the initialized word embeddings. Here $d_w$ denotes the dimension of word embeddings. $\mathbf{A}_w \in \mathbb{R}^{|\mathcal{V}_w|\times |\mathcal{V}_w|}$ represents the adjacency matrix derived from the co-occurrence statistics of words within the text, with each value determined by point-wise mutual information (PMI), denoted as $[\mathbf{A}_w]_{ij}=\text{max}(\text{PMI}(\mathcal{V}_{w,i}, \mathcal{V}_{w,j}), 0)$. The word graph extracts valuable semantic information from the perspective of words.

The POS graph $\mathcal{G}_p=\{\mathcal{V}_p,\mathbf{X}_p,\mathbf{A}_p\}$ comprises POS tags, such as nouns and adverbs, assigned to words, serving to disambiguate the syntactic roles of words. $\mathcal{V}_p$ represents the POS set and $\mathbf{X}_p\in \mathbb{R}^{|\mathcal{V}_p|\times d_p}$ is the initialized node features by one-hot encoding. Here, $d_p$ denotes the dimension of node features. $\mathbf{A}_p \in \mathbb{R}^{|\mathcal{V}_p|\times |\mathcal{V}_p|}$ denotes the adjacency matrix based on the co-occurrence statistics of POS tags within the text, with each element also specified by PMI, \textit{i.e.}, $[\mathbf{A}_p]_{ij}=\text{max}(\text{PMI}(\mathcal{V}_{p,i}, \mathcal{V}_{p,j}), 0)$. The POS graph mines valuable syntactic information from the perspective of POS tags.

The entity graph $\mathcal{G}_e=\{\mathcal{V}_e,\mathbf{X}_e,\mathbf{A}_e\}$ is composed of the entities presented in the NELL knowledge base. $\mathcal{V}_e$ denotes the set of entities and $\mathbf{X}_e \in \mathbb{R}^{|\mathcal{V}_e|\times d_e}$ denotes the initialized entity embeddings. Here, $d_e$ is the dimension of entity embeddings. $\mathbf{A}_e\in \mathbb{R}^{|\mathcal{V}_e|\times|\mathcal{V}_e|}$ denotes the adjacency matrix, where each element value is obtained from the cosine similarity between entity pairs, \textit{i.e.}, $[\mathbf{A}_e]_{ij}=\text{max}(\cos(\mathcal{V}_{e,i},\mathcal{V}_{e,j}),0)$. The entity graph captures auxiliary information from the perspective of entities. 
for better understanding.

\subsection{Multi-View Text Representation Learning}
After constructing the aforementioned three text-related component graphs $\mathcal{G}_\pi\!=\!\{\mathcal{V}_\pi, \mathbf{X}_\pi, \mathbf{A}_\pi\}, \pi \!\in\! \{w,e,p\}$ from different perspectives, we can individually apply graph convolution operations, as defined in Eq.\ref{gnn}, to obtain the high-quality node embeddings $\mathbf{H}_\pi\in\mathbb{R}^{|\mathcal{V}_\pi|\times d_\pi}$. In this way, we can fully leverage the interactions between nodes of the same type. 

Next, we construct view-specific adjacency matrices to establish connections between short texts and nodes of different types, namely words, POS tags, and entities, in order to obtain informative text representations under different views. Specifically, for the word graph $\mathcal{G}_w$ and POS graph $\mathcal{G}_p$, we utilize TF-IDF values to quantify the relationship between the text and nodes, represented as $\mathbf{T}_w\in\mathbb{R}^{N\times |\mathcal{V}_w|}$ and $\mathbf{T}_p\in\mathbb{R}^{N\times |\mathcal{V}_p|}$, where $N$ denotes the number of texts. For the entity graph $\mathcal{G}_e$ and its entity-specific matrix $\mathbf{T}_{e} \in \mathbb{R}^{N\times |\mathcal{V}_e|}$, if the $i$-th text contains the $j$-th entity, we set $\mathbf{T}_{e,ij}=1$; otherwise, $\mathbf{T}_{e,ij}=0$.

With the obtained $\mathbf{H}_\pi$ and $\mathbf{T}_\pi$, we can perform information aggregation operations to derive multi-view text representations, defined as follows:
\begin{equation}
\begin{aligned}
    \label{text}\mathbf{Z}_\pi&=\mathbf{T}_\pi\mathbf{H}_\pi, \pi\in\{w,e,p\},
\end{aligned}
\end{equation}
where $\mathbf{Z}_\pi \in \mathbb{R}^{N\times d_\pi}$. This operation can be viewed as an interpretation of each short text from the views of words, POS tags, and entities.

\subsection{Multi-View Contrastive Learning}
Unlike previous methods that require data augmentation operations to obtain augmented views, our approach naturally forms multi-view graphs during the graph construction process, resulting in corresponding multi-view text representations. Consequently, CL can be seamlessly executed. Moreover, existing methods exclusively engage in CL solely between the constructed pair of views, thereby disregarding other potential views, potentially resulting in the omission of crucial information. In contrast, our approach explicitly incorporates all available views. Specifically, we utilize a projection head $\Phi(\cdot)$ to map the obtained multi-view text representations $\mathbf{Z}_w, \mathbf{Z}_e$, and $\mathbf{Z}_p$ into the same hidden space, and then normalize these hidden embeddings to unit form, \textit{i.e.}, $\mathbf{P}_w=\text{norm}(\Phi(\mathbf{Z}_w)), \mathbf{P}_e=\text{norm}(\Phi(\mathbf{Z}_e))$, and $\mathbf{P}_p=\text{norm}(\Phi(\mathbf{Z}_p))$, so that they are comparable directly. For ease of presentation, we take the word view and the entity view as examples to perform CL.
\begin{equation}
\label{cl}
\begin{aligned}
        &\mathcal{L}_{w,i} = -\log\frac{\exp((\mathbf{P}_{w,i}\cdot \mathbf{P}_{e,i})/\tau)}{\mathbf{S}}, \\
        &\mathbf{S} = \sum_{k=1}^{N} [\mathbb{I}_{k\neq i}\exp((\mathbf{P}_{w,i} \cdot \mathbf{P}_{w,k})/\tau) + \\  &\qquad\exp((\mathbf{P}_{w,i} \cdot \mathbf{P}_{e,k})/\tau)], \\
            &\mathcal{L}_{w,e}=\frac{1}{2N}(\sum\nolimits_{i=1}^{N}\mathcal{L}_{w,i}+\sum\nolimits_{i=1}^{N}\mathcal{L}_{e,i}),
\end{aligned}
\end{equation}
where $\mathbf{P}_{w,i}$ and $\mathbf{P}_{e,i}$  are the representations of the same text under the word and entity views, which form the positive pair. $\tau$ and $\cdot$ represent the temperature parameter and dot product operator, respectively. $\mathbb{I}$ is the indicator function, equal to 1 if $k\neq i$, and 0 otherwise.

Similarly, we can perform the aforementioned process of CL on the word view and POS view of the text, as well as on the entity view and POS view of the text, to obtain contrastive losses $\mathcal{L}_{w,p}$ and $\mathcal{L}_{e,p}$, respectively. After that, we can obtain the final multi-view contrastive loss $\mathcal{L}_{cl}$, which can be defined as follows:
\begin{equation}
\label{ml_cl}
    \mathcal{L}_{cl} = \sum\nolimits_{i\in\{w,p,e\}, j\in\{w,p,e\},i\neq j}\mathcal{L}_{i,j}.
\end{equation}
\subsection{Model Optimization}
The original labeled short texts contain valuable supervised information, thus enabling effective utilization. Initially, we concatenate the multi-view text features, subsequently applying them to a linear classifier. The procedure can be expressed:
\begin{equation}
\label{ce}
\begin{aligned}
        \mathbf{Q}&=\mathbf{W}(\mathbf{Z}_w||\mathbf{Z}_p||\mathbf{Z}_e), \\
        \mathcal{L}_{ce}&=-\sum_{i\in \mathcal{D}_{lab}}\sum_j^c\mathcal{Y}_{ij}\log\mathbf{Q}_{ij},
\end{aligned}
\end{equation}
where $\mathcal{Y}$ denotes the ground-truth label and $\mathbf{W}$ denotes the trainable parameter. $c$ is the number of classes.

Finally, the adopted final loss can be denoted as:
\begin{equation}
\label{loss}
    \mathcal{L}=\mathcal{L}_{ce} + \mathcal{L}_{cl}.
\end{equation}

\begin{algorithm}
\caption{The Training of SimSTC}
\label{simstc}
\textbf{Input}: The evaluated corpus. \\
\textbf{Output}: Our proposed SimSTC. 
\begin{algorithmic}[1]
    \WHILE{\textit{not done}}
        \FOR{$\pi \in \{w,p,e\}$}{
        \STATE Construct the multi-view component graph $\mathcal{G}_\pi=\{\mathcal{V}_\pi,\mathbf{X}_\pi,\mathbf{A}_\pi\}$
        }
        \ENDFOR
        \STATE Update node embeddings for multi-view graphs using Eq.\ref{gnn}.
        \STATE Obtain multi-view text representations using Eq.\ref{text}.
        \STATE Perform multi-view CL using Eqs.\ref{cl} and \ref{ml_cl}.
        \STATE Conduct the cross-entropy loss using Eq.\ref{ce}.
        \STATE Optimize the model by the loss of Eq.\ref{loss}.
    \ENDWHILE
    \STATE \textbf{return}: The well-trained SimSTC.
\end{algorithmic}
\end{algorithm}

The training procedure of our framework can be found in Algorithm \ref{simstc}.

\subsection{Complexity Analysis}
In this section, we analyze the time complexity of the proposed model. In SimSTC, the most time-consuming operations are the GCN and CL operations. We perform 2-layer GCN operations on the constructed word graph $\mathcal{G}_w$, POS graph $\mathcal{G}_p$, and entity graph $\mathcal{G}_e$, with time complexities of $O(E_w(d_w+d)+2|\mathcal{V}_w|d^2)$, $O(E_p(d_p+d)+2|\mathcal{V}_p|d^2)$, and $O(E_e(d_e+d)+2|\mathcal{V}_e|d^2)$. Here, $d_w$, $d_p$, and $d_e$ represent the initial node dimensions of $\mathcal{G}_w$, $\mathcal{G}_p$, and $\mathcal{G}_e$, $d$ represents the hidden dimension, and $E_w$, 
$E_p$, and $E_e$ represent the number of non-zero elements in the adjacency matrices $\mathbf{A}_w$, $\mathbf{A}_p$, and $\mathbf{A}_e$ of $\mathcal{G}_w$, $\mathcal{G}_p$, and $\mathcal{G}_e$, respectively. Moreover, we conduct CL operations on the word-POS view, word-entity view, and POS-entity view. Since the number of texts is fixed at $N$, the time complexity for all CL is $O(Nd+N(2N-1)d)$, where the first term represent the time complexity of positive sample pairs, and the second term represents that of negative sample pairs. Therefore, the total time complexity is $O(E_w(d_w+d)+2|\mathcal{V}_w|d^2+E_p(d_p+d)+2|\mathcal{V}_p|d^2+E_e(d_e+d)+2|\mathcal{V}_e|d^2+3Nd+3N(2N-1)d)$, which can be simplified as $O(2|\mathcal{V}_w|d^2+2|\mathcal{V}_p|d^2+2|\mathcal{V}_e|d^2+3N(2N-1)d)$. This is acceptable to us.
\section{Theoretical Analysis}
In this section, we deeply explore the underlying principles of our framework from the perspective of mutual information. Specifically, we theoretically prove that SimSTC essentially increases the lower bound on mutual information for multi-view text embeddings, leading to improved performance. Next, we first present the following theorem.

\begin{theorem}
\label{theorem}
    Given the word-entity, word-POS, and entity-POS views, with corresponding text embeddings $\mathbf{Z}_w$, $\mathbf{Z}_e$, $\mathbf{Z}_p$, the mutual information between multi-view text embeddings satisfies the following inequality,
    \begin{equation}
    \label{mi}
        \sum_{i\in\{w,p,e\}, j\in \{w,p,e\},i \neq j} \mathrm{I}(\mathbf{Z}_i;\mathbf{Z}_j) \geq 3\log(N) - \mathcal{L}_{cl},
    \end{equation}
    where $\mathrm{I}(\cdot;\cdot)$ is the mutual information and $N$ is the number of datasets.
\end{theorem}
Theorem \ref{theorem} indicates that, under sufficient optimization of $\mathcal{L}_{cl}$, SimSTC tights the lower bound on mutual information, which is useful for downstream tasks. 
Moreover, intuitively, compared to our multi-view CL, dual-view CL may ignore crucial information. For instance, in the word-entity view, the mutual information between words and POS, or entities and POS, is entirely disregarded. 
\section{Experiment}
\subsection{Datasets}
To demonstrate the superiority of our proposed framework, we conduct thorough experiments on several benchmark datasets widely used in previous researches \cite{linmei2019heterogeneous,wang2021hierarchical}. The statistics of the datasets are shown in Table \ref{dataset}. Detailed descriptions of these datasets are provided below. (1) \textbf{Twitter} \cite{nlpbook} consists of a collection of crawled tweets provided by the NLTK Python package and is used for binary sentiment classification. (2) \textbf{MR} \cite{pang2005seeing} comprises numerous movie reviews labeled as positive or negative, and is also utilized for binary sentiment classification. (3) \textbf{Snippets} \cite{phan2008learning} has many web search snippets returned by the Google search engine. (4) \textbf{StackOverflow} \cite{xu2017self} consists of question titles extracted from the StackOverflow website.

\begin{table*}[ht]
\centering
\begin{tabular}{@{}c|ccccccc@{}}
\toprule
Dataset       & \#Doc  & \#Train(ratio) & \#Word & \#Entity & \#Tag & Avg.Length & \#Class \\ \midrule
Twitter       & 10,000 & 40 (0.40\%)    & 21,065 & 5,837    & 41    & 3.5        & 2       \\
MR            & 10,662 & 40 (0.38\%)    & 18,764 & 6,415    & 41    & 7.6        & 2       \\
Snippets      & 12,340 & 160 (1.30\%)   & 29,040 & 9,737    & 34    & 14.5       & 8       \\
StackOverflow & 20,000 & 400 (2\%)      & 2,632  & 3,229    & 42    & 8.3        & 20      \\ \bottomrule
\end{tabular}%
\caption{Statistics of evaluation datasets.}
\label{dataset}
\end{table*}

We adopt the same data preprocessing steps as previous studies: tokenizing each sentence and removing stop words and low-frequency words that appear fewer than five times in the corpus. Moreover, we randomly sample 40 labeled short texts from each category in the dataset, with half used for training, another half for validation, and the remaining data for testing the model performance, as suggested by previous studies \cite{linmei2019heterogeneous,wang2021hierarchical}.

\subsection{Baselines}
We mainly choose three types of baseline models for comparison to verify the effectiveness of the proposed SimSTC.
(I) \textit{Modern Deep Learning Models} contain \textbf{CNN} \cite{kim-2014-convolutional}, \textbf{LSTM} \cite{liu2015multi}, \textbf{BERT} \cite{devlin2018bert}, and \textbf{RoBERTa} \cite{liu2019roberta}. Here, we utilize the BERT-base and RoBERTa-base versions. (II) \textit{Graph-based models} consist of \textbf{TLGNN} \cite{huang2019text}, \textbf{HyperGAT} \cite{ding2020more}, \textbf{TextING} \cite{zhang2020every}, \textbf{DADGNN} \cite{liu2021deep}, and \textbf{TextGCN} \cite{yao2019graph}. (III) \textit{Deep Short Text Models} include \textbf{STCKA} \cite{chen2019deep}, \textbf{STGCN} \cite{ye2020document}, \textbf{HGAT} \cite{linmei2019heterogeneous}, \textbf{SHINE} \cite{wang2021hierarchical}, \textbf{NC-HGAT} \cite{su2022contrastive}, and \textbf{GIFT} \cite{liu2024improved}. Notably, we also provide several representative \textit{large language models (LLMs)}, comprising \textbf{GPT-3.5} \cite{ouyang2022training}, \textbf{Bloom-7.1B} \cite{workshop2022bloom}, \textbf{Llama2-7B} \cite{touvron2023llama}, \textbf{Llama3-8B} \cite{llama3modelcard}. 

\subsection{Implementation Details}
We employ 2-layer GCNs with 128 hidden units to encode the built multi-view graphs. 
we implement the projection head $\Phi(\cdot)$ by MLP with a 128-dimensional hidden layer. All the temperature parameters $\tau$ involved in multi-view CL are uniformly set to 0.5. We use the Adam to optimize the proposed model and set the learning rate to 0.001. We also adopt the early-stopping strategy, where if the loss does not decrease for 10 consecutive epochs on the validation set, the training is halted. We use the PyTorch library 1.10 to implement our model with Python 3.7. Moreover, we adopt the NVIDIA RTX 3090Ti GPU to accelerate the model training. For LLMs, we fine-tune GPT-3.5 using OpenAI's fine-tuning interface with the training data from the evaluation dataset and obtain results based on the prompts provided in Appendix. Additionally, for Bloom-7.1B, Llama2-7B, and Llama3-8B, we conduct comprehensive fine-tuning using the training data. To reduce GPU memory usage, we employ the Parameter Efficient Fine-Tuning (PEFT) method with LoRA and 4-bit quantization techniques provided by Hugging Face. The prompts used for STC in Bloom-7.1B, Llama2-7B, and Llama3-8B are the same as those used in GPT-3.5. 
\subsection{Evaluation Metric}
We utilize widely used accuracy (ACC) and macro-F1 score (F1) as evaluation metrics. All experiments are conducted five times, and the average results along with their corresponding standard deviations are reported to ensure statistical significance.

\begin{table*}[ht]
\centering
\renewcommand\arraystretch{1.15}
\resizebox{\textwidth}{!}{%
\begin{tabular}{@{}c|cc|cc|cc|cc@{}}
\toprule
\multirow{2}{*}{Model} & \multicolumn{2}{c|}{Twitter}    & \multicolumn{2}{c|}{MR}         & \multicolumn{2}{c|}{Snippets}   & \multicolumn{2}{c}{StackOverflow} \\ \cmidrule(l){2-9} 
                       & ACC            & F1             & ACC            & F1             & ACC            & F1             & ACC             & F1              \\ \midrule
CNN                    & 57.29$\pm$0.92          & 56.02$\pm$1.25          & 59.06$\pm$0.72          & 59.01$\pm$0.69          & 77.09$\pm$0.48          & 69.28$\pm$0.50          & 63.75$\pm$0.32           & 61.21$\pm$0.62           \\
LSTM                   & 60.28$\pm$0.70          & 60.22$\pm$0.79          & 60.89$\pm$0.58          & 60.70$\pm$0.72          & 75.89$\pm$0.52          & 67.72$\pm$0.42          & 61.62$\pm$0.63           & 60.49$\pm$0.61           \\
BERT                   & 54.92$\pm$0.28          & 51.16$\pm$0.35          & 51.69$\pm$0.52          & 50.65$\pm$0.36          & 79.31$\pm$0.53          & 78.47$\pm$0.30          & 66.94$\pm$0.29           & 67.26$\pm$0.32           \\
RoBERTa                & 56.02$\pm$0.39          & 52.29$\pm$0.19          & 52.55$\pm$0.26          & 51.30$\pm$0.25          & 79.55$\pm$0.19          & 79.02$\pm$0.22          & 69.91$\pm$0.34           & 70.35$\pm$0.36           \\ \midrule
TLGNN                  & 59.02$\pm$0.40          & 54.56$\pm$0.42          & 59.22$\pm$0.39          & 59.36$\pm$0.37          & 70.25$\pm$0.29          & 63.29$\pm$0.25          & 62.09$\pm$0.39           & 61.91$\pm$0.32           \\
HyperGAT               & 59.15$\pm$0.52          & 55.19$\pm$0.59          & 58.65$\pm$0.39          & 58.62$\pm$0.42          & 70.89$\pm$0.49          & 63.42$\pm$0.52          & 63.25$\pm$0.55           & 62.10$\pm$0.39           \\
TextING                & 59.62$\pm$0.72          & 59.22$\pm$0.79          & 58.89$\pm$0.62          & 58.76$\pm$0.69          & 71.10$\pm$0.53          & 70.65$\pm$0.59          & 65.37$\pm$0.71           & 64.63$\pm$0.77           \\
DADGNN                 & 59.51$\pm$0.39          & 55.32$\pm$0.49          & 58.92$\pm$0.32          & 58.86$\pm$0.21          & 71.65$\pm$0.39          & 70.66$\pm$0.36          & 66.26$\pm$1.09           & 65.10$\pm$1.03           \\
TextGCN                & 60.15$\pm$0.96          & 59.82$\pm$0.99          & 59.12$\pm$0.33          & 58.98$\pm$0.35          & 77.82$\pm$0.47          & 71.95$\pm$0.35          & 67.02$\pm$0.51           & 66.51$\pm$0.39           \\ \midrule
STCKA                  & 57.56$\pm$0.26          & 57.02$\pm$0.25          & 53.25$\pm$0.31          & 51.19$\pm$0.33          & 68.96$\pm$0.41          & 61.27$\pm$0.39          & 59.72$\pm$0.29           & 59.65$\pm$0.21           \\
STGCN                  & 64.33$\pm$0.55          & 64.29$\pm$0.72          & 58.25$\pm$0.71          & 58.22$\pm$0.51          & 70.01$\pm$0.62          & 69.93$\pm$0.40          & 69.23$\pm$0.19           & 69.10$\pm$0.15           \\
HGAT                   & 63.21$\pm$0.37          & 57.02$\pm$0.42          & 62.75$\pm$0.61          & 62.36$\pm$0.70          & 82.36$\pm$0.39          & 74.44$\pm$0.42          & 67.35$\pm$0.29           & 66.92$\pm$0.32           \\
SHINE                  & 72.54$\pm$0.39          & 72.19$\pm$0.32          & 64.58$\pm$0.39          & 63.89$\pm$0.35          & 82.39$\pm$0.61          & 81.62$\pm$0.75    & 73.05$\pm$0.50           & 72.73$\pm$0.39           \\
NC-HGAT                & 63.76$\pm$0.25          & 62.94$\pm$0.23          & 62.46$\pm$0.29          & 62.14$\pm$0.27          & 82.42$\pm$0.32    & 74.62$\pm$0.31          & 67.59$\pm$0.39           & 67.02$\pm$0.41           \\ 
GIFT                  & \underline{73.16$\pm$0.66} & \underline{73.16$\pm$0.72} & \underline{65.21$\pm$0.36} & \underline{65.16$\pm$0.39} & \underline{83.73$\pm$0.52} & \underline{82.35$\pm$0.56} & \underline{83.07$\pm$0.36} &
\underline{82.94$\pm$0.39} \\
\midrule
SimSTC             &    \textbf{73.51$\pm$0.32}     &     \textbf{73.49$\pm$0.35}   &       \textbf{66.52$\pm$0.49} &         \textbf{66.36$\pm$0.39}          & \textbf{85.14$\pm$0.72} & \textbf{83.94$\pm$0.65} & \textbf{84.07$\pm$0.26}  & \textbf{83.98$\pm$0.25} \\ \midrule
\textcolor{gray}{GPT-3.5}                & \textcolor{gray}{81.23$\pm$0.39}          & \textcolor{gray}{80.02$\pm$0.19}          & \textcolor{gray}{87.43$\pm$0.95} & \textcolor{gray}{86.62$\pm$0.92}          & \textcolor{gray}{66.50$\pm$0.72}          & \textcolor{gray}{63.48$\pm$0.79}          & \textcolor{gray}{81.29$\pm$0.94}           & \textcolor{gray}{81.16$\pm$0.82}           \\
\textcolor{gray}{Bloom-7.1B}               & \textcolor{gray}{87.52$\pm$0.75} & \textcolor{gray}{86.56$\pm$0.66} & \textcolor{gray}{87.03$\pm$0.71}          & \textcolor{gray}{86.96$\pm$0.79} & \textcolor{gray}{71.39$\pm$0.59}          & \textcolor{gray}{60.76$\pm$0.62}          & \textcolor{gray}{81.42$\pm$0.55}           & \textcolor{gray}{81.65$\pm$0.51}           \\
\textcolor{gray}{Llama2-7B}             & \textcolor{gray}{87.45$\pm$0.29}  & \textcolor{gray}{86.43$\pm$0.22}    & \textcolor{gray}{87.26$\pm$0.39}    & \textcolor{gray}{86.69$\pm$0.27}    & \textcolor{gray}{73.05$\pm$0.52}          & \textcolor{gray}{68.11$\pm$0.55}          & \textcolor{gray}{82.29$\pm$0.50}     & \textcolor{gray}{81.96$\pm$0.49}      
  \\ 
\textcolor{gray}{Llama3-8B}             & \textcolor{gray}{87.65$\pm$0.26}  & \textcolor{gray}{86.55$\pm$0.29}    & \textcolor{gray}{87.36$\pm$0.29}    & \textcolor{gray}{86.72$\pm$0.22}    & \textcolor{gray}{75.25$\pm$0.56}          & \textcolor{gray}{70.16$\pm$0.26}          & \textcolor{gray}{82.99$\pm$0.55}     & \textcolor{gray}{82.26$\pm$0.36}      
  \\\bottomrule
\end{tabular}%
}
\caption{Results (\%) of the accuracy and macro-F1 score with standard deviation on several short text datasets. We highlight the best performance in bold excluding LLMs based on the pairwise t-test with 95\% confidence.}
\label{res}
\end{table*}
\section{Result}

\begin{table*}[ht]
\centering
\resizebox{0.99\textwidth}{!}{%
\begin{tabular}{@{}ccc|cccccccc@{}}
\toprule
\multirow{2}{*}{Word-POS View} & \multirow{2}{*}{POS-Entity View} & \multirow{2}{*}{Word-Entity View} & \multicolumn{2}{c}{Twitter}     & \multicolumn{2}{c}{MR}          & \multicolumn{2}{c}{Snippets}    & \multicolumn{2}{c}{StackOverflow} \\ \cmidrule(l){4-11} 
                               &                                  &                                   & ACC            & F1             & ACC            & F1             & ACC            & F1             & ACC             & F1              \\ \midrule
-                              & -                                & -                                 & 70.60          & 69.74          & 61.94          & 61.60          & 79.27          & 78.10          & 80.29           & 80.12           \\
\checkmark                              & -                                & -                                 & 72.59          & 72.55          & 64.18          & 64.18          & 83.63          & 82.74          & 82.26           & 82.13           \\
-                              & \checkmark                                & -                                 & 71.92          & 71.86          & 62.99          & 62.46          & 82.96          & 81.59          & 82.15           & 81.92           \\
-                              & -                                & \checkmark                                 & 72.56          & 72.50          & 63.95          & 63.72          & 83.66          & 82.98          & 82.20           & 82.09           \\
\checkmark                              & \checkmark                                & -                                 & 72.90          & 72.62          & 65.20          & 65.12          & 84.52          & 83.09          & 82.36           & 82.22           \\
\checkmark                              & -                                & \checkmark                                 & 73.15          & 72.76          & 65.76          & 65.72          & 84.66          & 83.07          & 82.55           & 82.50           \\
-                              & \checkmark                                & \checkmark                                 & 73.29          & 72.86          & 65.29          & 65.22          & 84.79          & 83.16          & 82.49           & 82.42           \\
\checkmark                              & \checkmark                                & \checkmark                                 & \textbf{73.51} & \textbf{73.49} & \textbf{66.52} & \textbf{66.36} & \textbf{85.14} & \textbf{83.94} & \textbf{84.07}  & \textbf{83.98}  \\ \bottomrule
\end{tabular}%
}
\caption{Ablation results of different model variants.}
\label{ablation}
\end{table*}

\subsection{Model Performance}
We present a comprehensive comparison of our proposed model with other baseline models, showcasing the results in Table \ref{res}. Based on the quantified results, we have the following in-depth analysis and observation.

\noindent $\bullet$ We observe that our model achieves competitive performance on all the selected evaluated datasets compared to the first three types of baselines. Moreover, it even significantly outperforms currently popular LLMs on several datasets, such as Snippets and StackOverflow, which demonstrates its effectiveness in STC tasks. One crucial contributing factor is the construction of multi-view graphs, which facilitate a comprehensive exploration of semantic and syntactic knowledge within the corpus. This approach also enables the incorporation of auxiliary knowledge from external knowledge graphs, effectively addressing the issue of semantic sparsity in short texts. Moreover, the introduction of multi-view CL, without compromising the original text semantics, facilitates the acquisition of discriminative text features from numerous unlabeled texts, thereby mitigating the severe issue of label scarce.

\noindent $\bullet$ We find that on the MR and Twitter datasets, LLMs perform considerably surpass other models. One plausible reason is that LLMs with numerous parameters have superior text understanding, making binary sentiment classification tasks extremely simple for them. Additionally, it is difficult to determine whether these LLMs have been exposed to the test set during the large-scale unsupervised pre-training, which could lead to data leakage issues, especially for closed-source models like GPT-3.5. However, on the snippets and StackOverflow datasets, they do not outperform our model. We speculate that the increase in text categories makes the classification task more challenging. Moreover, these domain-specific texts may account for a small proportion in the pre-training corpus, leading to its inefficiency. 
Although our model does not perform as well as the LLMs on the Twitter and MR datasets, it still outperforms other types of baselines.

\noindent $\bullet$ We find that graph-based deep short text models (excluding STCKA) outperform both modern deep learning models and graph-based models in various datasets, which is in accordance with our expectations. A reasonable explanation is that previous deep short text models are specifically designed to address the semantic sparsity of short texts and the scarcity of labels, thus mitigating the negative impact of the key issues. In contrast, modern deep learning models and graph-based models rely solely on their original feature extraction capabilities, which may not fully leverage their advantages in short text classification applications.

\subsection{Ablation Study}

To validate the effectiveness of the introduced multi-view CL, we investigate the impact of using various combined views of short texts on the model performance. Here, when any contrastive view is not introduced, it means that we simply input the merged multi-view textual features $\mathbf{Z}_w||\mathbf{Z}_p||\mathbf{Z}_e$ into a linear classifier. According to Table \ref{ablation}, we clearly observe that when compared to the original model without any contrastive views, the addition of any view proves to be beneficial for the model. Each contrastive view can provide the model with crucial and diverse information that is previously overlooked by other models. For example, in the MR dataset, the Word-POS contrastive view brings the 2.24\% absolute improvement in model accuracy compared to the base model without any contrastive one. Moreover, as we introduce more views, the model performance continues to improve, aligning with expectations, as it enables the model to capture much useful information for downstream tasks.
For example, the model variant with Word-POS and POS-Entity views achieves the 2.30\% higher accuracy on the Twitter dataset compared to the base model.

\begin{table}[ht]
\centering
\resizebox{0.46\textwidth}{!}{%
\begin{tabular}{@{}c|cccc@{}}
\toprule
Dataset    & Twitter   & MR      & Snippets & StackOverflow \\ \midrule
Bloom      & \multicolumn{4}{c}{31,457,280}                 \\ \cmidrule(l){2-5} 
Llama2     & \multicolumn{4}{c}{33,554,432}                 \\ \cmidrule(l){2-5}
Llama3     & \multicolumn{4}{c}{33,554,432}                 \\ \cmidrule(l){2-5}
SimSTC     & 1,577,818 & 664,538 & 773,608  & 490,116       \\ \cmidrule(l){2-5} 
ratio (\%) & 4.70      & 1.98    & 2.31     & 1.46          \\ \bottomrule
\end{tabular}%
}
\caption{The trainable parameters of SimSTC are compared to those of LLMs. The last row shows the ratio of trained parameters in our model to those in Llama3.}
\label{params_ratio}
\end{table}


\subsection{Model Efficiency}
We demonstrate the efficiency of our model by comparing it with LLMs in terms of trainable parameters. 
According to Table \ref{params_ratio}, we find that SimsTC has far fewer trainable parameters than LLMs, comprising at most 5\% of the total trainable parameters in LLMs. However, our model still considerably outperforms these LLMs on certain datasets, such as Snippets, demonstrating its effectiveness for this focused task. 
Moreover, this phenomenon suggests that our model achieves an excellent balance between efficiency and performance.


\subsection{Discussion}
In the era of prevalent LLMs, an intriguing question emerges: is there a continued necessity for the development of superior text embedding models? Through the research expounded in this paper, we proffer an affirmative response. Given that contemporary large models are trained primarily on text generation tasks, specifically predicting the subsequent token, as opposed to assessing the comprehensive quality of sentence representation, the direct utilization of large models for text representation fails to yield the anticipated results. Consequently, the pursuit of an exceptional text representation model retains significant relevance.
\section{Conclusion}
In this work, we develop a simple graph contrastive framework, coined SimSTC, for STC tasks. We do not perform any explicit data augmentation to generate contrastive views. On the contrary, we fully leverage the naturally produced different text views to obtain informative multi-view text representations when constructing multi-view graphs for CL. 
We conduct extensive experiments on the evaluated datasets. Compared to other competitive models, SimSTC has demonstrated satisfactory performance, even significantly outperforming currently popular LLMs on several datasets. 


%
\begin{links}
    \link{Code}{https://github.com/KEAML-JLU/SimSTC}
\end{links}

\section*{Acknowledgments}
This work is supported in part by funds from the National Key Research and Development Program of China (No. 2021YFF1201200), the National Natural Science Foundation of China (No. 62172187 and No. 62372209). Fausto Giunchiglia’s work is funded by European Union’s Horizon 2020 FET Proactive Project (No. 823783).
\bibliography{aaai25}

\end{document}